\documentclass[letterpaper]{article} % DO NOT CHANGE THIS
\usepackage{aaai25}
\usepackage{times}  % DO NOT CHANGE THIS
\usepackage{helvet}  % DO NOT CHANGE THIS
\usepackage{courier}  % DO NOT CHANGE THIS
\usepackage[hyphens]{url}  % DO NOT CHANGE THIS
\usepackage{graphicx} % DO NOT CHANGE THIS
\usepackage{natbib}  % DO NOT CHANGE THIS AND DO NOT ADD ANY OPTIONS TO IT
\usepackage{caption} % DO NOT CHANGE THIS AND DO NOT ADD ANY OPTIONS TO IT
\usepackage{algorithm}
\usepackage{algpseudocode}

\usepackage{times}
\usepackage{soul}
\usepackage{url}
\usepackage[utf8]{inputenc}
\usepackage{amsmath}
\usepackage{amsthm}
\usepackage{booktabs}
\usepackage[switch]{lineno}
\usepackage{pifont}

\usepackage{algorithm}
\usepackage{algpseudocode} 
\usepackage{amsmath}       
\usepackage{graphicx}      
\usepackage{caption}       
\usepackage{multirow}
\usepackage[most]{tcolorbox}

\usepackage{newfloat}
\usepackage{listings}
\DeclareCaptionStyle{ruled}{labelfont=normalfont,labelsep=colon,strut=off} % DO NOT CHANGE THIS
\floatstyle{ruled}
\newfloat{listing}{tb}{lst}{}
\floatname{listing}{Listing}
\title{ValuesRAG: Enhancing Cultural Alignment Through Retrieval-Augmented Contextual Learning}
\author{
Wonduk Seo$^{1,2}$ \hspace{0.5em}
Zonghao Yuan$^3$\hspace{0.5em}
Yi Bu$^{1,4}$\thanks{Yi Bu is the corresponding author for this work.} \\
}

\affiliations{
$^1$Peking University \hspace{0.5em}
$^2$Enhans \hspace{0.5em}
$^3$Tsinghua University\\
$^4$Peking University Chongqing Research Institute of Big Data\\
$^1$\{seowonduk, buyi\}@pku.edu.cn, $^2$yzh23@mails.tsinghua.edu.cn
}

\usepackage{bibentry}
\begin{document}

\maketitle

\begin{abstract}
Ensuring cultural values alignment in Large Language Models (LLMs) remains a critical challenge, as these models often embed Western-centric biases from their training data, leading to misrepresentations and fairness concerns in cross-cultural applications. Existing approaches—such as role assignment and few-shot learning—struggle to address these limitations effectively due to their reliance on pre-trained knowledge, limited scalability, and inability to capture nuanced cultural values. To address these issues, we propose \emph{\textbf{ValuesRAG}}, a novel and effective framework that applies Retrieval-Augmented Generation (RAG) with In-Context Learning (ICL) to integrate cultural and demographic knowledge dynamically during text generation. Leveraging the World Values Survey (WVS) dataset, ValuesRAG first generates summaries of values for each individual. We subsequently curate several representative regional datasets to serve as test datasets and retrieve relevant summaries of values based on demographic features, followed by a reranking step to select the top-k relevant summaries. We evaluate ValuesRAG using $6$ diverse regional datasets and show that it consistently outperforms baselines: including zero-shot, role-assignment, few-shot, and hybrid methods, both in main experiments and ablation settings. Notably, ValuesRAG achieves the best overall performance over prior methods, demonstrating its effectiveness in fostering culturally aligned and inclusive AI systems. Our findings underscore the potential of dynamic retrieval-based methods to bridge the gap between global LLM capabilities and localized cultural values.
\end{abstract}

\begin{figure}
\centering
\includegraphics[width=1\linewidth]{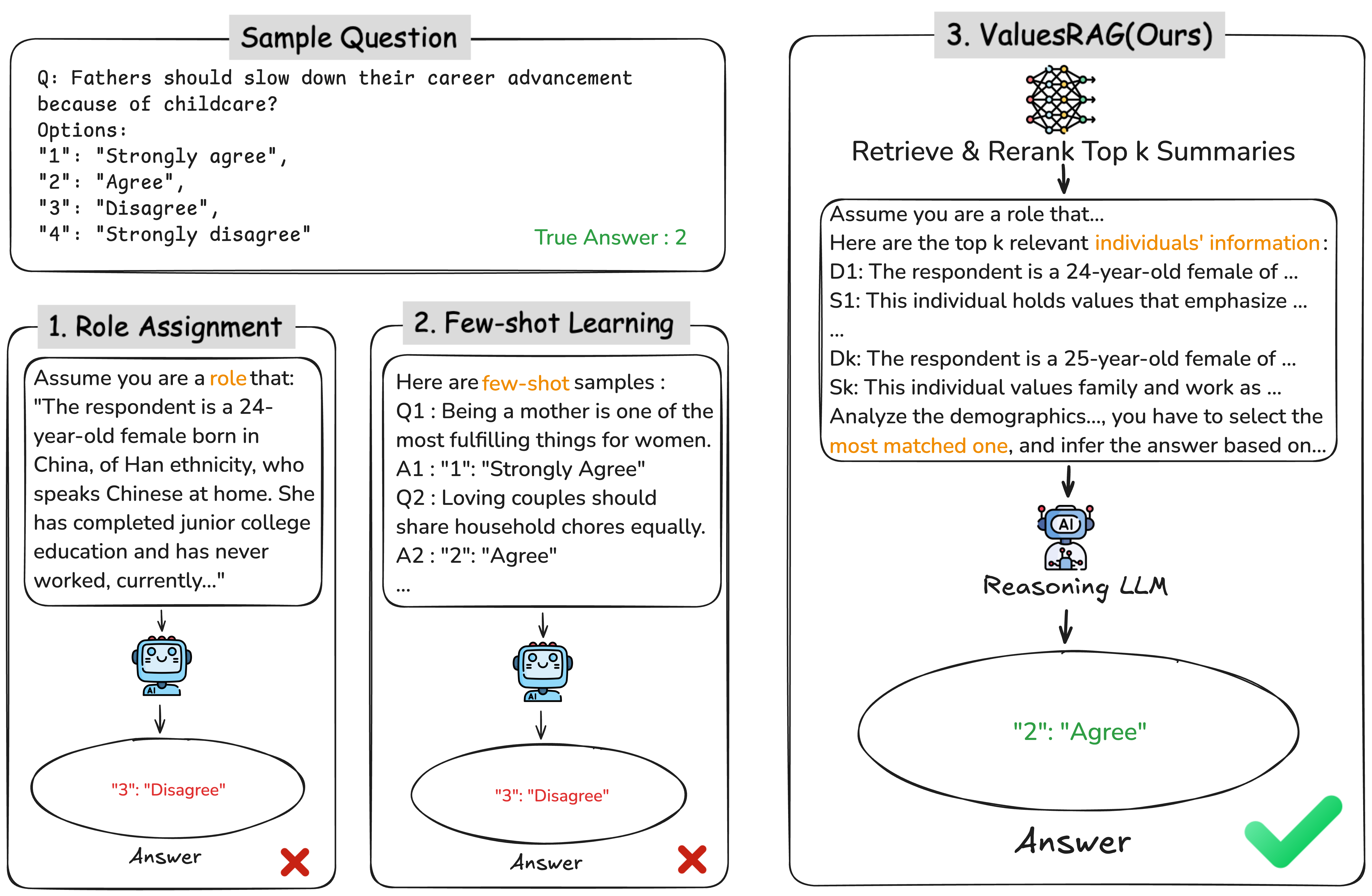}
\caption{\textbf{Overview of different approaches for cultural alignment.} Comparing two baseline methods, namely Role Assignment and Few-Shot Learning, and our proposed ValuesRAG framework.}
\label{fig:Overview_baseline}
\end{figure}

\section{Introduction}
The rapid advancement of Large Language Models (LLMs) has revealed pressing challenges in cultural values alignment
~\cite{singhGlobalMMLUUnderstanding2024,kharchenkoHowWellLLMs2024,hu2024generative}. Predominantly trained on Western data sources~\cite{achiam2023gpt,touvron2023llama,jiang2023mistral}, LLMs inherently reflect Western cultural norms and social biases, raising concerns about their applicability in global contexts. These biases present significant challenges when deploying LLMs in cross-cultural environments, often resulting in misrepresentations and stereotypical outputs~\cite{gallegosBiasFairnessLarge2024,xieCanLargeLanguage2024,potterHiddenPersuadersLLMs2024,huangDemocratizingValueAlignment2024}. 
Despite ongoing efforts to address these issues, existing strategies often fall short. While some countries have developed localized LLMs, such as China’s ERNIE~\cite{sun2021ernie}, ChatGLM~\cite{glm2024chatglm}, DeepSeek~\cite{liu2024deepseek}, and South Korea’s HyperCLOVA~\cite{yoo2024hyperclova}, these models also exhibit biases inherited from their respective training datasets. As a result, cultural and social biases embedded in LLMs remain a critical concern, compelling researchers to explore more robust frameworks for cultural alignment~\cite{gallegosBiasFairnessLarge2024,xieCanLargeLanguage2024,potterHiddenPersuadersLLMs2024}.

Recent studies have proposed several approaches, such as \emph{role-assignment} approaches~\cite{10.1093/pnasnexus/pgae346} and \emph{few-shot learning} techniques ~\cite{choenniSelfAlignmentImprovingAlignment2024}, to mitigate these cultural biases. However, these methods still face several challenges: (1) Role-assignment approaches, relying solely on the model’s pre-trained knowledge, provide pre-defined demographic information but fail to incorporate explicit values alignment text, which subsequently introduces stereotypes and biases rooted in Western-centric training data; (2) While offering example-based guidance, few-shot learning methods struggle to comprehensively capture the complex cultural values due to the limited correlation between different values dimensions, thus remain ineffective on values-related tasks that differ significantly from the examples; (3) In addition, these methods can only align with the values of a single individual, and singular values cannot represent the universal values of individuals with similar characteristics. 

To address these challenges, we propose \emph{\textbf{ValuesRAG}}, a novel framework that utilizes Retrieval-Augmented Generation (RAG) and In-Context Learning (ICL) to dynamically incorporate cultural knowledge during text generation (see Figure \ref{fig:Overview_baseline}). Our framework leverages the World Values Survey (WVS) dataset~\cite{haerpfer2022world}, a globally recognized and comprehensive dataset that explores values across countries using rigorous social science methodologies.
First, we specifically generate summaries for each topic, followed by generating individuals' summaries of values and demographic profiles in parallel. After constructing the knowledge base, we retrieve the top-$100$ relevant summaries based on demographic features, followed by a reranking step to ensure the most relevant top-\textit{k} summaries are selected. Finally, we utilize a reasoning LLM that filters the most relevant demographic profiles and applies reasoning grounded in the retrieved values to generate the final answer to the question.

We evaluate the performance of ValuesRAG by comparing it against several baseline approaches, including: (1) \emph{zero-shot inference}, (2) \emph{role-assignment-only method}~\cite{10.1093/pnasnexus/pgae346}, (3) \emph{few-shot learning}~\cite{choenniSelfAlignmentImprovingAlignment2024}, and (4) \emph{a hybrid method combining both (1) and (2)}. To ensure a comprehensive evaluation, we curated diverse regional survey QA datasets which are designed to capture values-related question-answer pairs. Extensive experimental results show significant improvements in cultural and contextual understanding, demonstrating that ValuesRAG outperforms the baselines. Compared to previous methods that heavily depend on pre-trained knowledge or limited demonstrations, ValuesRAG dynamically retrieves and integrates multiple similar individual values summaries based on demographic features, enabling richer value representations and more context-aware responses compared to approaches relying on a single predefined prompt or role. 

In addition, ablation studies on varying the number of retrieved summaries and on using only value-augmented generation confirm ValuesRAG’s robust performance under different configurations. Adjusting the number of retrieved documents shows that moderate retrievals can balance diversity and relevance, also maintaining high accuracy across multiple benchmarks. Meanwhile, ValuesRAG surpasses the baselines through purely values-based generation.

These findings highlight ValuesRAG’s potential to foster inclusive AI systems, enhancing the reliability and fairness of AI-driven applications. Our study demonstrates ValuesRAG's robust capabilities on a global scale, also suggesting its applicability in aligning the values of diverse groups within a single country. ValuesRAG provides a cost-efficient tool for public policymakers and scientists from various disciplines to refine social simulations, enabling more precise predictions of policy outcomes~\cite{li2024political}. This, in turn, facilitates the creation of fairer and more effective policies. Moreover, NGOs can leverage ValuesRAG to develop LLMs that reflect specific value orientations while maintaining strong alignment with users' values, thereby increasing their persuasive impact. This approach benefits the promotion and spread of values that contribute to the planet's sustainable development and the long-term well-being of human society.

\section{Related Work}

\subsection{Evaluation of LLMs' Cultural Bias}
Pre-trained models are facing growing criticism for their inherent social biases, with cultural bias emerging as a particularly nuanced and pervasive issue~\cite{10.1093/pnasnexus/pgae346}. While safety concerns and social discrimination in language models are typically explicit and well-recognized~\cite{liuTrustworthyLLMsSurvey2024}, cultural biases often manifest subtly, reflecting dominant cultural perspectives embedded within training data. Studies have shown that LLMs often exhibit cultural biases aligned with the values of developed countries, resulting in the under-representation of perspectives from less developed regions ~\cite{manviLargeLanguageModels2024,durmusMeasuringRepresentationSubjective2024}. This imbalance not only perpetuates existing cultural hierarchies but also limits the global applicability of these models ~\cite{manviLargeLanguageModels2024}. Various benchmarks and evaluation methods have been proposed to assess the cultural biases of pre-trained models~\cite{gallegosBiasFairnessLarge2024}. For instance, 
Caliskan et al.~\shortcite{caliskanSemanticsDerivedAutomatically2017} pioneered the use of word embeddings as quantitative measures of bias, while Webster et al. ~\shortcite{websterMeasuringReducingGendered2021} developed probability-based metrics to evaluate gender bias embedded in pre-trained models. More recently, Karinshak et al.~\shortcite{karinshakLLMGLOBEBenchmarkEvaluating2024} introduced \emph{LLM-GLOBE}, a benchmark where LLMs generate both quantitative and open-ended answers to values assessment questions, with subsequent evaluation using the LLM-as-a-Jury Protocol. These evaluation methods collectively highlight the complex nature of cultural bias in LLMs and the need for multifaceted assessment approaches.

\begin{table*}[htbp]
    \centering
    % \small
    \begin{tabular}{@{}lllllll@{}}
    \toprule
    \textbf{Category} & \textbf{Dataset} & \textbf{Abbreviation} & \textbf{Region} & \textbf{Year} & \textbf{N} & \textbf{VQ} \\ 
    \midrule
    \multirow{1}{*}{\emph{\textbf{Retrieval Corpus}}} 
    & World Values Survey 
    & WVS 
    & Global 
    & 2017--2022
    & 97.2k & 259  \\
    \midrule
    \multirow{6}{*}{\emph{\textbf{Test Datasets}}} 
    & European Values Study 
    & EVS 
    & Europe 
    & 2017
    & 59.4k & 211 \\
    & The General Social Survey 
    & GSS 
    & North America 
    & 2021--2022
    & 8.2k & 44 \\
    & Chinese General Social Survey 
    & CGSS 
    & East Asia 
    & 2021
    & 8.1k & 58 \\
    & India Survey Dataset 
    & ISD
    & South Asia 
    & 2019--2020
    & 30.0k & 33 \\
    & AmericasBarometer 
    & LAPOP 
    & Latin America 
    & 2021
    & 59.1k & 48 \\
    & Afrobarometer 
    & Afrobarometer 
    & Africa 
    & 2022
    & 48.1k & 144 \\
    \bottomrule
    \end{tabular}
    \caption{\textbf{Overview of the datasets utilized in our study.} The \emph{Retrieval Corpus} (WVS) includes global data collected between 2017 and 2022, providing the basis for generating cultural summaries of values and validation for our method. The \emph{Test Datasets} consist of six region-specific surveys, each capturing socio-cultural information from distinct geographic areas and time frames. N represents sample size in thousands (k). VQ represents the number of values-related questions.}
    \label{tab:datasets-used}
\end{table*}

\subsection{Mitigation of LLMs' Cultural Bias}

Techniques such as Reinforcement Learning from Human Feedback (RLHF)~\cite{shenLargeLanguageModel2023,jiAIAlignmentComprehensive2024} are commonly employed for aligning LLMs with human values. However, these single-dimensional alignment methods are insufficient for mitigating cultural bias, as cultural values are inherently diverse, dynamic, and context-dependent, varying significantly across different regions and societies~\cite{huangDemocratizingValueAlignment2024}. Addressing cultural biases has become a critical area of research, with various strategies being proposed to enhance cultural sensitivity in LLMs. For instance, Tao et al. ~\shortcite{10.1093/pnasnexus/pgae346} adopted national and cultural role assignments to adjust the cultural values of LLMs, while Masoud et al.~\shortcite{masoudllm} developed a soft prompt tuning approach to mitigate bias. Moreover, Choenni and Shutova~\shortcite{choenniSelfAlignmentImprovingAlignment2024} employed few-shot in-context learning to align cultural behaviors, demonstrating promising results in specific contexts. However, these approaches face significant limitations in fully capturing the complexity of cultural alignment. Tao et al.’s approach~\cite{10.1093/pnasnexus/pgae346} mainly depends on national and cultural roles without explicitly integrating values assignments, causing an overreliance on latent internal representations. Meanwhile, Choenni and Shutova’s few-shot learning approach~\cite{choenniSelfAlignmentImprovingAlignment2024} similarly falls short of modeling cultural alignment in all its complexity. We therefore use these methods as baselines to benchmark our proposed approach.

\section{Datasets}
In this section, we first introduce the World Values Survey (WVS) as our retrieval corpus, highlighting its extensive coverage, global representativeness, and relevance for values-related studies. 
Subsequently, we describe six regional test datasets, which are carefully selected to ensure geographic, cultural, and demographic diversity. 

\subsection{Retrieval Corpus} 

WVS\footnote{https://www.worldvaluessurvey.org/wvs.jsp}~\cite{haerpfer2022world} is a globally recognized dataset that investigates human beliefs, values, and cultural norms through structured surveys conducted across multiple countries. WVS is selected as our retrieval corpus especially due to its numerous advantages: 
\begin{description}  
    \item[\emph{1. Broad recognition and inclusiveness:}]  
    WVS is widely recognized and frequently used by governments, social scientists, and major international organizations in comparative values studies. It currently covers 120 countries, representing 94.5\% of the global population, ensuring broad geographic and cultural representation.  
    \item[\emph{2. Expert-designed and accessible:}]  
    The dataset is meticulously designed by leading domain experts to conduct comprehensive surveys of values, ensuring reliability, rigor, and relevance. It is publicly accessible, enabling reproducibility and transparency in research.  
    \item[\emph{3. Effective structure and large scale:}]  
    WVS has well-organized and comprehensive demographic questions, making it effective for retrieval tasks. Its large sample size (97,221 respondents) is also suitable for RAG tasks.  
\end{description}  

Since values evolve gradually over time, WVS is conducted in waves, with each wave occurring every five years. For our study, we utilize the most recent wave, spanning from 2017 to 2022.

The WVS codebook includes over 600 indicators, with 259 values-related and 31 demographic-related questions. The value questions span 13 topics, such as social trust, post-materialism, and political interest, as shown in Table \ref{tab:wvs_topics}. It covers most dimensions of values, allowing for a comprehensive and accurate measure of each respondent's values. We randomly select 20\% (52 questions) per topic for validation and use the remaining 80\% (207 questions) for summary generation. The 31 demographic features, including country, sex, age, education, social class, and employment status, are used to generate demographic summaries for retrieval tasks.

\begin{figure}[htbp]
\includegraphics[width=1\linewidth]{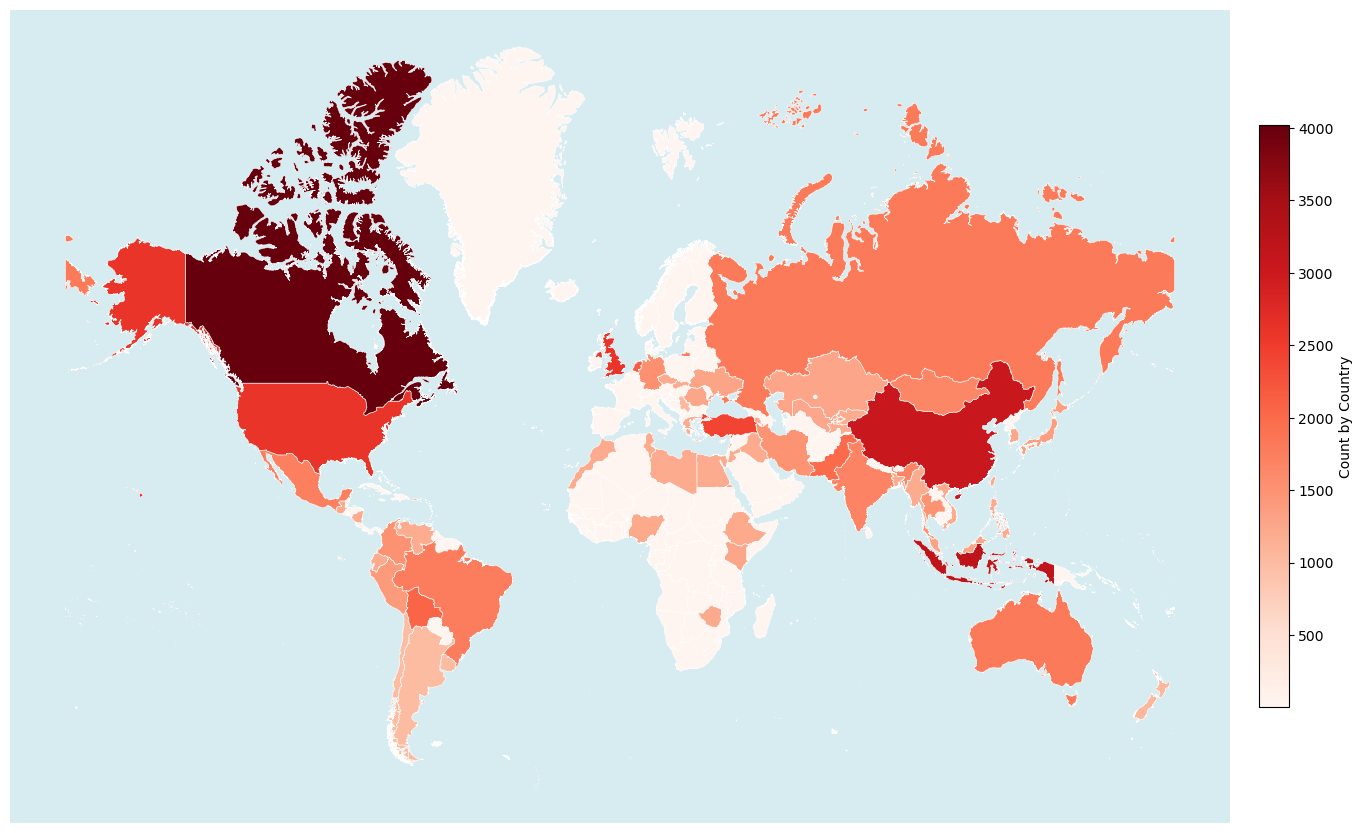}
    \caption{\textbf{Distribution of countries in the WVS dataset.} The WVS dataset covers surveys conducted in over $120$ countries across all major regions, providing broad geographic and demographic representation for building a reliable retrieval corpus in RAG-based frameworks.}
    \label{fig:wvs}
\end{figure}

\begin{table}[htbp]
\centering
\begin{tabular}{@{}l r@{}}
\toprule
\textbf{Topic} & \textbf{Count} \\
\midrule
Social Values, Norms, Stereotypes                     & 45 \\
Happiness and Wellbeing                               & 11 \\
Social Capital, Trust and Organizational Membership   & 47 \\
Economic Values                                       & 6  \\
Perceptions of Corruption                             & 9  \\
Perceptions of Migration                              & 10 \\
Perceptions of Security                               & 21 \\
Index of Postmaterialism                              & 6  \\
Perceptions about Science and Technology              & 6  \\
Religious Values                                      & 12 \\
Ethical Values                                        & 23 \\
Political Interest and Political Participation        & 35 \\
Political Culture and Political Regimes               & 25 \\
\bottomrule
\end{tabular}
\caption{\textbf{Distribution of Values-related Questions in WVS.} The questions were categorized into 13 topics with a total of 259 questions covering most of the dimensions of values}
\label{tab:wvs_topics}
\end{table}

\subsection{Test Datasets} 
We select six regional surveys to serve as test datasets based on the following criteria:
\begin{description}  
    \item[\emph{1. Demographic and values coverage:}]  
    The datasets provide demographic features closely aligned with WVS's questions, along with sufficient values-related features to enable meaningful comparisons and analyses.  
    \item[\emph{2. Temporal proximity:}]  
    The datasets exhibit close temporal proximity to WVS Wave 7 (2017–2022), thereby allowing aligned comparisons and ensuring thorough consistency across diverse global evaluations.  
\end{description}  

The regions in our test datasets are meticulously chosen to encompass a wide range of geographic, cultural, and demographic diversity, ensuring that the data accurately reflects the majority of the global population. All of them are publicly accessible and are statistically representative at national or regional levels, which guarantees their reliability and validity. Also, these test sets include both values-related questions and demographic characteristics. The demographic characteristics are used to generate summaries, serving as retrieval targets for RAG. The values-related questions are utilized as test questions to calculate accuracy (ACC).\footnote{A detailed description of the method for computing accuracy (ACC) is provided in~\emph{Section Experiments - Evaluation Metrics}.} The specific datasets used in our evaluation, along with their characteristics, are described below:

\paragraph{EVS}The first dataset comes from the \textit{European Values Study}\footnote{https://europeanvaluesstudy.eu}, a large-scale, cross-national, and longitudinal survey research program designed to explore values, beliefs, and attitudes across Europe. This dataset includes a total of 211 values-related questions and captures 34 demographic characteristics of the respondents. We select EVS 2017, conducted in 2017, ensuring alignment with the World Values Survey (WVS) in terms of the timeframe. 

\paragraph{GSS}We select GSS to represent the population of the United States. \textit{The General Social Survey}\footnote{https://gss.norc.org} is a sociological survey that has been conducted since 1972 by the National Opinion Research Center (NORC) at the University of Chicago. Its primary purpose is to collect and analyze data on the opinions, behaviors, and demographic characteristics of adults in the United States, thereby monitoring societal change and the growing complexity of American society. Its questionnaire covers a comprehensive and wide range of topics, including many values-related questions. Specifically, within the GSS, we identify 44 questions as values-related and 33 questions as demographic characteristics.

\paragraph{CGSS} \textit{The Chinese General Social Survey}\footnote{http://cgss.ruc.edu.cn}, initiated in 2003, is China's earliest national, comprehensive, and continuous academic survey project. Conducted by the National Survey Research Center at Renmin University of China, the CGSS systematically collects data at multiple levels, including society, communities, families, and individuals. CGSS only provided the questionnaire and data in Chinese, which we have translated into English to ensure its usability. We ultimately compile a total of 58 values-related questions and 13 demographic characteristics.

\paragraph{ISD}To ensure that our experiment covers as much of the world's population as possible, we made efforts to include India within the scope of our test set. However, we were unable to obtain data from several government surveys in India, thus we used data published by the Pew Research Center instead. The Pew Research Center's \textit{India Survey Dataset}\footnote{https://www.pewresearch.org/dataset/india-survey-dataset} is a comprehensive resource that captures the perspectives of 29,999 Indian adults on various aspects of society, including religious beliefs and practices, identity, nationalism, and societal tolerance. Conducted through face-to-face interviews between November 17, 2019, and March 23, 2020, the survey encompassed participants from diverse religious backgrounds, such as Hindus, Muslims, Sikhs, Christians, Buddhists, Jains, and others. This dataset covers 33 values-related questions and 23 demographic characteristics.

\paragraph{LAPOP}The Latin American Public Opinion Project (LAPOP)\footnote{https://www.vanderbilt.edu/lapop} is a research institute based at Vanderbilt University in Nashville, Tennessee. LAPOP's most notable survey is the \textit{AmericasBarometer}, the most extensive survey of democratic public opinion and behavior covering the Americas, including North, Central, South America, and the Caribbean. This survey measures democratic values and behaviors through voter surveys, providing valuable insights into public sentiments across the region. We select this dataset to represent the population of Latin America. There are 48 values-related questions and 12 demographic characteristics.

\paragraph{Africa}\textit{Afrobarometer}\footnote{https://www.afrobarometer.org} is a pan-African, non-partisan research network established in 1999 that conducts public attitude surveys on democracy, governance, economic conditions, and related issues across Africa. We selected data from the 8th round of Afrobarometer (collected in 2022), covering 34 African countries. After screening, a total of 144 values-related questions and 14 demographic characteristics are obtained.

% To be specific, we select the European Values Study~\cite{evsEuropeanValuesStudy2022} as the representative dataset for Europe, as it is the largest values survey in the region. For the United States, we select the General Social Survey~\cite{Davern2024}, which is the most comprehensive social survey in the country. The Chinese General Social Survey~\cite{Bian01102012} serves as the representative dataset for China due to its comprehensive sampling methodology and scientific rigor. For India, where national survey data were largely inaccessible during our study, we use Pew Research Center’s survey data~\cite{SahgalEvans2021} to represent the Indian population. The AmericasBarometer~\cite{lapop2021}, conducted by the LAPOP Lab, is selected to represent Latin American countries, as it covers 32 countries across the region. Finally, Afrobarometer~\cite{afrobarometer2023} is chosen as the representative dataset for Africa. A detailed summary of these datasets is presented in \autoref{tab:datasets-used}.

\begin{figure*}[!htpb]
    \centering
    \includegraphics[width=1.0\textwidth]{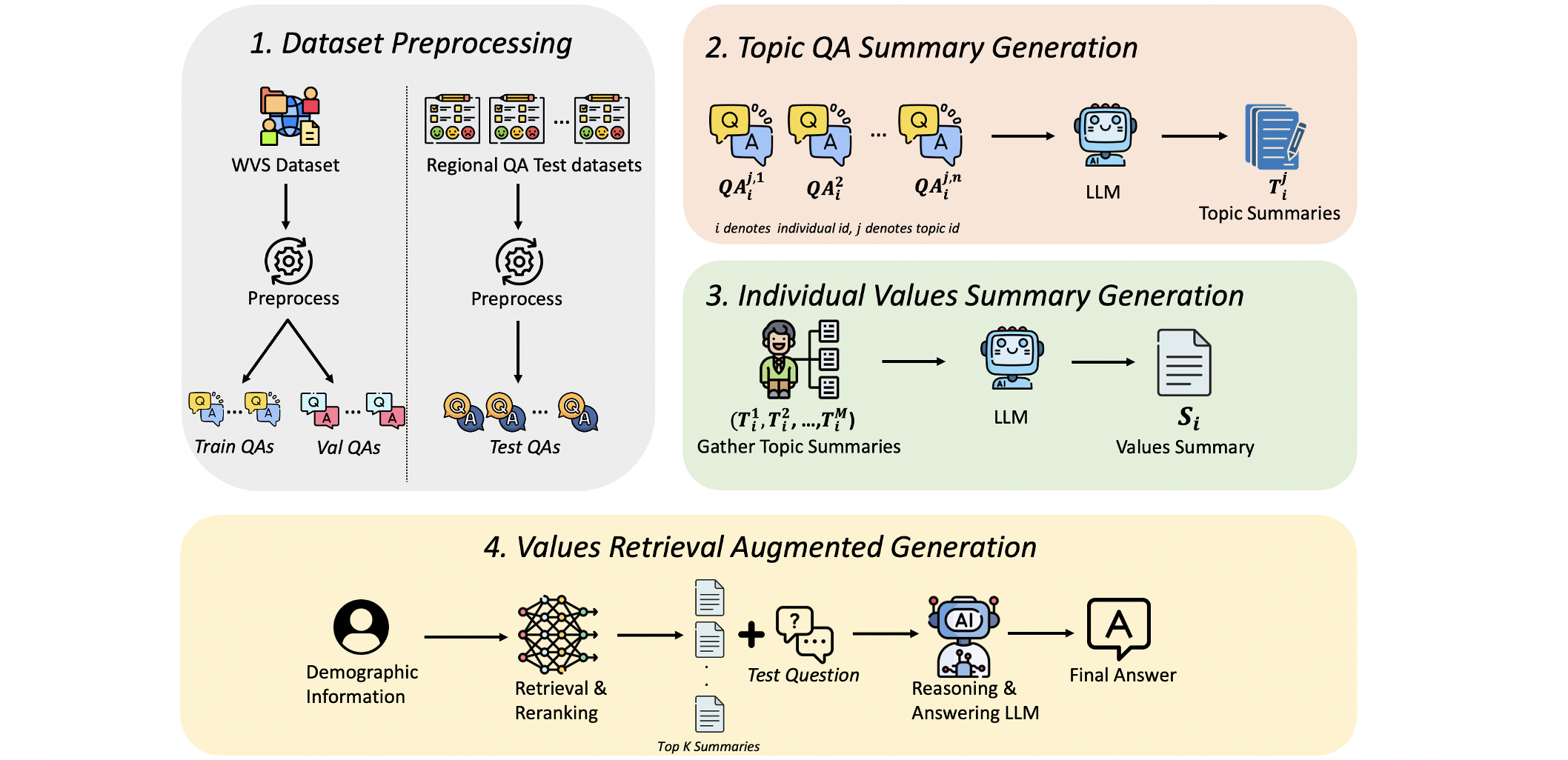}
    \caption{\textbf{Overview of the proposed ValuesRAG framework for cultural alignment.} The framework comprises four key stages: (1) Dataset preprocessing to separate training and test QA pairs from WVS and regional datasets, (2) Topic-wise summary generation using LLMs for each individual, (3) Aggregation of topic summaries into comprehensive individual values profiles, and (4) Retrieval-Augmented Generation that retrieves and reranks relevant value summaries based on demographic similarity to guide final response generation.}
    \label{fig:values_rag}
\end{figure*}

\section{Methodology}
In this section, we present the \emph{ValuesRAG} which is specifically designed to address cultural biases and enhance contextual alignment in LLM-driven scenarios through a Retrieval-Augmented Generation (RAG) approach. \emph{ValuesRAG} consists of three key components: (1) \emph{Values and Demographic Summary Generation}, which extracts and summarizes cultural values and demographic information from large-scale datasets; (2) \emph{Values-Augmented Generation}, which incorporates these summaries into the generative process to align responses with the cultural context; and (3) \emph{Retrieval-based Values Alignment}, which dynamically assigns relevant individual values to queries based on demographic profiles. An overview of the \emph{ValuesRAG} framework is provided in Figure~\ref{fig:values_rag}.

\subsection{Values and Demographic Summary Generation}
To systematically generate concise summaries of values and demographics for each individual, we process the dataset in three stages. First, the dataset is stratified by topics and split into train and validation sets, ensuring that the distribution of each topic is preserved, as described in previous section. In parallel, topic-based summaries and demographic summaries are generated separately. For topic-based summaries, values-related QA sets are used to produce summaries for each topic, while demographic summaries are generated using demographic-related QA sets:
\begin{equation}
\begin{aligned}
T_i^j &= f_{\text{gen}}(\text{QA}_i^{j, 1}, \text{QA}_i^{j, 2}, \dots, \text{QA}_i^{j, N_j}), \\
D_i &= f_{\text{gen}}(\text{QA}_i^{\text{demo}, 1}, \text{QA}_i^{\text{demo}, 2}, \dots, \text{QA}_i^{\text{demo}, K}).
\end{aligned}
\end{equation}
where \( f_{\text{gen}} \) denotes the generative model, \(T_i^j\) is the summary for topic \(j\) of individual \(i\), based on \(N_j\) values-related QA pairs, and \(D_i\) represents the demographic summary derived from \(K\) demographic-related QA pairs. Finally, individual summaries are constructed by combining all topic summaries:
\begin{equation}
S_i = f_{\text{gen}}(T_i^1, T_i^2, \dots, T_i^M),
\end{equation}
with \(M\) denoting the total number of topics. The result, denoted as \(S_i\), forms a comprehensive values summary for individual \(i\). These generate summaries serve as structured references for retrieval in later stages and are used to augment the validation set for evaluation as well.

\subsection{Values Augmented Generation}
Once the comprehensive summaries for each individual are generated in the previous step, we construct an augmented generation process for evaluating on the validation question-answer data. For each validation question, we concatenate the corresponding individual’s values summary with the question itself, forming a context-rich input for the LLM:

\begin{equation}
C_i = \text{concatenate}\left(S_i, \text{Q}_i^{\text{val}, k}\right)
\end{equation}
where \(C_i\) represents the combined context, \(S_i\) is the values summary for individual \(i\), and \(\text{Q}_i^{\text{val}, k}\) is the \(k\)-th validation question. We subsequently concatenate \(C_i\) with the demographic summary \(D_i\) to further enhance the context, enabling the generation of responses based on both values and demographic information:
\begin{equation}
A_i = f_{\text{gen}}\left(C_i, D_i\right)
\end{equation}
Here, \(A_i\) represents the answer generated by the function \(f\), and \((C_i, D_i)\) embeds the augmented context \(C_i\) and demographic information \(D_i\) into a structured input format. Additionally, we utilize Chain-of-Thought (CoT) prompting~\cite{wei2022chain} to enhance reasoning and emulate the behavior of the corresponding individual, ensuring responses that are contextually aligned with the values captured in the summaries and demographic characteristics.

\subsection{Retrieval-based Values Alignment}
To dynamically assign relevant values to test individuals, we leverage demographic information as documents for retrieval. The demographic data from both the train (retrieval corpus) and test datasets are preprocessed into a structured context format, as described earlier, and embeddings are generated for each demographic context using a sentence-transformer-based model~\cite{reimers2019sentence}.\footnote{A detailed description of the model used is provided in ~\emph{Section Experiments - Models Used}.} We first retrieve the top-100 most similar summaries of values for each test individual by computing the cosine similarity between the embeddings of the test and train demographics:

\begin{equation}\small
\text{Sim}(E_{\text{test}}, E_{\text{train}}) = \frac{E_{\text{test}} \cdot E_{\text{train}}}{\|E_{\text{test}}\| \|E_{\text{train}}\|}
\end{equation}

$E_{\text{test}}$ and $E_{\text{train}}$ specifically represent the embeddings of the test and training (retrieval corpus) demographic contexts, respectively, and $\text{Sim}(\cdot,\cdot)$ denotes the cosine similarity score. The top-100 embeddings with the highest similarity scores are initially selected as candidates. We subsequently apply a reranking step to refine the selection and identify the most relevant summaries among the retrieved candidates.

The reranking process evaluates the semantic relevance by passing each candidate summary embedding $E_{C_j}$ along with the test individual's embedding $E_{\text{test}}$ through a neural reranker $f_{\text{rerank}}$, which outputs a relevance score $s_j$:

\begin{equation}
s_j = f_{\text{rerank}}\left(E_{\text{test}}, E_{C_j}\right), \quad j \in \{1, 2, \dots, 100\}
\end{equation}

We then sort the candidate summaries based on these scores in descending order and select the top-$k$ summaries with the highest scores as the final reranked set $R'_k$:

\begin{equation}
R'_k = \text{Top-}k\left(\{C_j\}_{j=1}^{100}, \{s_j\}_{j=1}^{100}\right)
\end{equation}

The reranked top-$k$ summaries $R'_k$ are incorporated into the prompts, enriching the contextual alignment of the generated responses. In detail, for each test individual, the retrieved and reranked summaries are combined into the final prompt, and the answer is subsequently generated using the function $f_{\text{gen}}$:

\begin{equation}
\begin{aligned}
P_{\text{test}} &= \left(D_{\text{test}}, R'_1, R'_2, \dots, R'_K, Q_{\text{test}}\right), \\
A_{\text{test}} &= f_{\text{gen}}\left(P_{\text{test}}\right).
\end{aligned}
\end{equation}
Here, \(P_{\text{test}}\) is the final prompt, \(D_{\text{test}}\) is the demographic information of the test individual, \(\{R'_1, R'_2, \dots, R'_K\}\) represents the top-\(k\) reranked summaries, and \(Q_{\text{test}}\) is the test question. \(A_{\text{test}}\) denotes the generated answer for the test question, and \(f_{\text{gen}}\) represents the generation function.

This retrieval-based approach, followed by reranking, enhances reasoning by explicitly guiding the LLM to critically evaluate which retrieved values best align with the test individual’s demographic characteristics. The final prompts are then used to generate answers following the chain-of-thought prompting strategy, ensuring that the responses are contextually coherent and culturally aligned with the test individual’s profile.  For the comprehensive implementation of the \emph{ValuesRAG}, we provide the \textbf{Algorithm~\ref{algo:valuesrag}}, which systematically outlines the processes of values and demographic summary generation, values-augmented generation, and retrieval-based values alignment, as shown below:

\begin{algorithm}
\caption{\emph{Values Generation and Retrieval Process}}\label{algo:valuesrag}
\begin{algorithmic}[1]
\Require Dataset $\mathcal{D}$ with topics and demographic QA pairs, Generative Model $f_{\text{gen}}$, Embedding Model $f_{\text{embed}}$, Reranking Model $f_{\text{rerank}}$, Retrieval Top-$K$

\Statex \textit{// Values and Demographic Summary Generation}
\For{each individual $i$ in $\mathcal{D}$}
    \State Generate topic-based values summaries $T_i^j$ for each topic $j$
    \State Generate demographic summaries $D_i$ 
    \State Combine $T_i^j$ into comprehensive values summary $S_i$
\EndFor

\Statex \textit{// Values Augmented Generation}
\For{each validation question $Q_i^{\text{val}, k}$ of individual $i$}
    \State Construct context $C_i = \text{concat}(S_i, Q_i^{\text{val}, k})$
    \State Augment context with demographic summary $D_i$
    \State Generate answer $A_i = f_{\text{gen}}(C_i, D_i)$
\EndFor

\Statex \textit{// Retrieval-based Values Alignment}
\For{each test individual $i$}
    \State Compute embeddings $E_{\text{test}} = f_{\text{embed}}(D_{\text{test}})$
    \State Retrieve top-100 values summaries by similarity:
    \Statex \hspace{\algorithmicindent} $\text{Sim}(E_{\text{test}}, E_{\text{train}})$
    \State Rerank top-$K$ summaries:
    \Statex \hspace{\algorithmicindent} $R'_k = f_{\text{rerank}}(E_{\text{test}}, E_{C_j})$
    \State Final prompt $P_{\text{test}} = (D_{\text{test}}, R'_1, \dots, R'_K, Q_{\text{test}})$
    \State Generate answer $A_{\text{test}} = f_{\text{gen}}(P_{\text{test}})$
\EndFor
\end{algorithmic}
\end{algorithm}

\section{Experiments}
\subsection{Setup}

\paragraph{Models Used.}We utilize \emph{GPT-4o-mini}~\cite{achiam2023gpt} and \emph{Gemini-1.5-Flash}~\cite{team2024gemini} for our generation tasks, which are accessed via APIs. We set the temperature parameter of these models to $0.7$ to achieve a balance between coherence and creativity. For the retrieval task, we employed the \emph{E5 (base)} model~\cite{wang2022text}, which generates embeddings and retrieves the top 100 most relevant summaries of values based on cosine similarity. 

To refine this retrieval, we apply a reranker using the \emph{GTE-multilingual-reranker-base} model~\cite{zhang2024mgtegeneralizedlongcontexttext}. Each retrieved candidate is paired with the test individual’s demographic summary and passed as a sentence pair to the reranker. The model outputs a relevance score for each pair, which we use to rank all candidates. The top-k summaries with the highest scores are then selected as the final set used for generation. This two-stage process: retrieval followed by reranking, ensures both semantic similarity and contextual relevance.

% ################################################################ 
\paragraph{Prompts Used}
We provide the prompts designed for various components of the ValuesRAG, including prompts for performing question answering tasks, as well as for generating values and demographic summaries below:

\begin{tcolorbox}[colback=white!20,colframe=darkgray!80,title=Prompt for Question Answering, halign=flush left]
\textbf{Task:}  
Respond to the question as the target individual, selecting the answer that aligns with their values and demographic context.

\textbf{Rules:}
\begin{itemize}
    \item Step-by-step analysis using retrieved demographics and values data.
    \item Maintain the target individual's perspective throughout the analysis.
    \item Provide the response in JSON format, without additional explanation.
\end{itemize}

\textbf{Steps for Inferring:}
\begin{enumerate}
    \item Analyze the demographics (age, gender, cultural background, social class, religion, and economic class) of retrieved individuals. Compare them with the target individual.
    \item Identify individuals whose demographics most closely match the target individual. Note their IDs.
    \item Based on the matched individual's values, infer how the target would respond.
    \item Select the response that best aligns with the inferred values, and return only the integer representing the selected option.
\end{enumerate}
\end{tcolorbox}

\begin{tcolorbox}[colback=white!20,colframe=darkgray!80,title=Prompt for Values Summary Generation]
You are a summarization expert with expertise in extracting key insights from complex data. Based on the provided context, summarize this person's values in one paragraph.
\end{tcolorbox}

\begin{tcolorbox}[colback=white!20,colframe=darkgray!80,title=Prompt for Demographic Summary Generation]
You are a summarization expert with expertise in extracting key insights from complex data. Based on the provided context, summarize this person's demographics in one paragraph.
\end{tcolorbox}
% ################################################################ 

\paragraph{Baseline Methods and Implementation.}
Our baseline methods include: (1) \emph{Zero-shot inference}, (2) \emph{the role-assignment-only approach}~\cite{10.1093/pnasnexus/pgae346}, (3) \emph{a few-shot learning method}~\cite{choenniSelfAlignmentImprovingAlignment2024}, and (4) \emph{a hybrid method that combines both (1) and (2)}. For the role-assignment baseline, we specifically use the same demographic summaries as in ValuesRAG to ensure fairness by assigning roles based on demographic information from the survey data. For the few-shot method, we follow the approach outlined in the previous work, where we randomly select five examples from the test set as prompts. The hybrid method combines both strategies, assigning roles based on demographic summaries and augmenting the prompts with five randomly selected few-shot examples from the test set.
Additionally, We use \emph{ValuesRAG} with $k=3$ retrieved summaries—chosen to provide a good balance between retrieval diversity and contextual relevance.\footnote{A detailed analysis of varying $k$ and its implications on model performance is provided in \emph{Ablation Study - Varying the Number of Retrieved Summaries}.}

\paragraph{Evaluation Metrics.}
We utilize accuracy as the primary evaluation metric, following previous work~\cite{choenniSelfAlignmentImprovingAlignment2024}, by converting multiple-choice responses into a binary format for consistency and simplicity. Specifically, we transform each original response $r_i$, ranging on a scale (e.g., Likert scale from 1 to 10), into a binary value $b_i$ indicating disagreement or agreement as follows. Here, $m$ denotes the midpoint of the response scale:

\begin{equation}
b_i =
\begin{cases}
0, & \text{if } r_i \leq m \quad \text{(disagree)} \\[6pt]
1, & \text{if } r_i > m \quad \text{(agree)}
\end{cases}
\end{equation}

This categorization metrics effectively captures distinct answer patterns and aligns naturally with values-related questions often posed through Likert-scale formats.

\begin{table*}[t]
\centering
\resizebox{\textwidth}{!}{%
\begin{tabular}{@{}c l c c c c c c c@{}}
\toprule[1pt]
\textbf{Model} & \textbf{Methods} & \textbf{EVS} & \textbf{GSS} & \textbf{CGSS} & \textbf{ISD} & \textbf{LAPOP} & \textbf{Africa} & \textbf{Avg. Accuracy} \\
\midrule

%=========================
% gpt-4o-mini
%=========================
\multirow{5}{*}{\centering \textbf{GPT-4o mini}}
& Zero-shot Inference
    & 0.5566 & 0.6026 & 0.4019 & 0.6109 & 0.4195 & 0.3923 & 0.4973 \\

& Role-Assignment~(\citeyear{10.1093/pnasnexus/pgae346})
    & 0.5738 & \underline{0.7564} & 0.4813 & 0.6164 & \underline{0.4742} & \underline{0.5563} & \underline{0.5764} \\

& Few-Shot Learning~(\citeyear{choenniSelfAlignmentImprovingAlignment2024})
    & 0.5271 & 0.6538 & 0.4631 & 0.5804 & 0.4220 & 0.4258 & 0.5120 \\

& Hybrid Method
    & \underline{0.5938} & 0.7292 & \underline{0.5048} & \underline{0.6330} & 0.4414 & 0.5305 & 0.5721 \\

& \textbf{ValuesRAG}\textsuperscript{\textdagger}
    & \textbf{0.6021}\textsuperscript{*} 
    & \textbf{0.7781}\textsuperscript{*} 
    & \textbf{0.5387}\textsuperscript{*} 
    & \textbf{0.7001}\textsuperscript{*} 
    & \textbf{0.5030}\textsuperscript{*} 
    & \textbf{0.5953}\textsuperscript{*} 
    & \textbf{0.6195}\textsuperscript{*} \\
\midrule

%=========================
% gemini
%=========================
\multirow{5}{*}{\centering \textbf{Gemini 1.5 Flash}}
& Zero-shot Inference 
    & 0.5419
    & 0.6408
    & 0.4502
    & 0.6017
    & 0.4149
    & 0.4181
    & 0.5113\\
& Role-Assignment ~(\citeyear{10.1093/pnasnexus/pgae346})
    & 0.5598
    & \underline{0.7493}
    & 0.4770
    & 0.6048
    & \textbf{0.4747}
    & 0.5262
    & 0.5653\\
& Few-Shot Learning ~(\citeyear{choenniSelfAlignmentImprovingAlignment2024})
    & 0.5225
    & 0.6376
    & 0.4559
    & 0.5782
    & 0.4194
    & 0.4758
    & 0.5149\\
& Hybrid Method
    & \underline{0.5845}
    & 0.7193
    & \underline{0.5026}
    & \underline{0.6253}
    & 0.4448
    & 0.5166
    & \underline{0.5655}\\
& \textbf{ValuesRAG}\textsuperscript{\textdagger} 
    & \textbf{0.5869} % not significant
    & \textbf{0.7686}\textsuperscript{*} 
    & \textbf{0.5337}\textsuperscript{*} 
    & \textbf{0.6789}\textsuperscript{*}
    & \underline{0.4705}
    & \textbf{0.5473}\textsuperscript{*}
    & \textbf{0.5977}\textsuperscript{*}\\
\bottomrule[1pt]
\end{tabular}%
}% end resizebox
\caption{\textbf{Accuracy scores for various methods compared with multiple baselines across six regional datasets.} \(k\) indicates the number of summaries to be retrieved. 
\textbf{Bold text} indicates the best performance, 
\underline{underlined text} indicates the second-best performance.
\textsuperscript{*} denotes significant improvements (paired \(t\)-test with Holm-Bonferroni correction, \(p<0.05\)) over all baseline model(s).
\textsuperscript{\textdagger} denotes our proposed method.}
\label{tab:accuracy_comparison}
\end{table*}

\begin{table*}[htbp]
% \small
\centering
\begin{tabular}{@{}c c c c c c c c c@{}}
\toprule[1pt]
\textbf{Model} & \textbf{Num(K)} & \textbf{EVS} & \textbf{GSS} & \textbf{CGSS} & \textbf{ISD} & \textbf{LAPOP} & \textbf{Africa} & \textbf{Avg. Accuracy} \\
\midrule

\multirow{4}{*}{\centering \textbf{GPT-4o mini}}
& 1 & 0.5960 & \underline{0.7722} & \underline{0.5347} & 0.6853 & 0.4682 & 0.5905 & 0.6078 \\
& \textbf{3} & \underline{0.6021} & \textbf{0.7781} & \textbf{0.5387} & 0.7001 & 0.5030 & \textbf{0.5953} & \textbf{0.6195} \\
& 5 & \textbf{0.6052} & 0.7706 & 0.5301 & \textbf{0.7016} & \textbf{0.5061} & 0.5905 & \underline{0.6174} \\
& 10 & 0.6020 & 0.7380 & 0.5317 & \underline{0.7014} & \underline{0.5030} & 0.5680 & 0.6074 \\
\midrule

\multirow{4}{*}{\centering \textbf{Gemini 1.5 Flash}}
& 1 & 0.5753 & 0.7668 & 0.5272 & 0.6646 & 0.4548 & 0.5369 & 0.5876 \\
& \textbf{3} & \textbf{0.5869} & \underline{0.7686} & \textbf{0.5337} & \textbf{0.6789} & \textbf{0.4705} & \underline{0.5473} & \textbf{0.5977} \\
& 5 & \underline{0.5868} & \textbf{0.7690} & \underline{0.5303} & 0.6734 & \underline{0.4661} & \textbf{0.5498} & \underline{0.5959} \\
& 10 & 0.5852 & 0.7665 & 0.5279 & \underline{0.6773} & 0.4509 & 0.5464 & 0.5924 \\

\bottomrule[1pt]
\end{tabular}%

\caption{\textbf{Accuracy scores across six regional datasets for the ablation study:~\emph{Varying the Number of Retrieved Summaries}}. Num($K$) (\(k \in \{1, 3, 5, 10\}\)) indicates the number of demographic summaries retrieved.
\textbf{Bold text} indicates the best performance, 
\underline{underlined text} indicates the second-best performance.}
\label{tab:k_ablation}
\end{table*}

\subsection{Experimental Analysis}

In our experiments, we compare \emph{ValuesRAG} with four baseline methods: (1) zero-shot inference, (2) role-assignment, (3) few-shot learning, and (4) a hybrid approach that combines role-assignment and few-shot learning. Following previous work, as detailed in the Evaluation Metrics section, we specifically evaluate model performance using accuracy. All responses are binarized based on contrasting answer patterns to ensure consistency across different values-related questions. Experimental results are shown in Table~\ref{tab:accuracy_comparison}. 

We find that the role-assignment method generally surpasses both zero-shot and few-shot approaches. By grounding the agent’s responses in a clearly defined demographic context, it ensures more consistent performance. Yet, role assignment can sometimes lead to overly narrow representations when demograhpic roles are interpreted stereotypically. Meanwhile, few-shot learning can incorporate example-driven context, but its limited number of prompts may not consistently address the intricate ways individuals' belifs diverge within similar social settings. As a result, it struggles to generalize to the multifaceted nature of human values, particularly when faced with unexpected or complex cultural scenarios. The hybrid method, which merges role assignment and few-shot prompts, does offer a partial improvement in contextual diversity. However, it still remains insufficient for capturing the full spectrum of nuances that can arise from overlapping demographic factors and idiosyncratic personal persepectives.

In contrast, ValuesRAG overcomes these challenges by dynamically retrieving and integrating specific values-related cultural data for each agent. By focusing on values as the primary retrieval targets, this retrieval-augmented framework enables the model to include an expansive set of contextual clues, helping it reflect the depth and breadth of each individual's background and values. Crucially, our ValuesRAG provides a more adaptive mechanism for representing the subtle interplay of personal beliefs and cultural norms via avoiding the limitations of rigid demographic labels or small-sample prompts. ValuesRAG more effectively captures the complex dynamics that can shape a respondent's stance on different questions. Evaluations across diverse test datasets demonstrate that ValuesRAG with \(k=3\) consistently outperforms baseline methods, highlighting its ability to better represent cultural diversity, improve contextual alignment, and enhance overall model performance.

\subsection{Ablation Study}
\label{sec:ablation_main}
We conduct two ablation studies to analyze the configuration and robustness of ValuesRAG. First,  we vary the number of retrieved summaries (\(k\)) to examine how retrieval depth affects the model’s performance. We subsequently isolate the effect of using only values-based generation.

\subsubsection{Varying the Number of Retrieved Summaries}
\label{sec:ablation_k_values}

To quantify how the number of retrieved summaries \(k\) impacts \emph{ValuesRAG}'s performance, we evaluate \(k \in \{1,3,5,10\}\). Table~\ref{tab:k_ablation} reports accuracy on six regional datasets for both GPT-4o-mini and Gemini 1.5 Flash. 

For GPT-4o-mini, increasing \(k\) from 1 to 3 boosts accuracy on five of six datasets, rising from an average of 0.6078 to 0.6195, and yields the best scores on GSS (0.7781), CGSS (0.5387), and Africa (0.5953). Although \(k=5\) further improves EVS (0.6052) and ISD (0.7016), it degrades GSS and CGSS, dropping the average to 0.6174. At \(k=10\), performance falls across most datasets (average 0.6074), indicating that an excessive number of summaries dilute relevance. Similarly, for Gemini 1.5 Flash, \(k=3\) attains the highest overall accuracy (0.5977), with top scores on EVS (0.5869), CGSS (0.5337), ISD (0.6789), and LAPOP (0.4705). While \(k=5\) slightly edges out \(k=3\) on GSS (0.7690) and Africa (0.5498), it lowers EVS and ISD and yields a lower average (0.5959). Retrieval at \(k=10\) further declines.

These results reveal a clear trade-off: \(k=1\) constrains diversity, whereas \(k>3\) introduces marginally relevant or noisy summaries and increases latency. Consequently, \(k=3\) achieves the best balance between contextual breadth, accuracy gains, and computational cost. We thus adopt \(k=3\) as our default retrieval depth across all experiments.

\subsubsection{Impact of Values-Only Generation}
\label{sec:ablation_values_only}

To validate the robustness of ValuesRAG, we perform an ablation study using only values context augmented generation, thereby excluding the impact of demographic summaries on the model’s performance. We use the WVS validation set: separated from the training data, which served as the retrieval corpus (as outlined in Section:~\emph{Datasets}), to evaluate the models. Table \ref{tab:ablation_performance} presents a comparison of our method, using \emph{only summaries of values}, against four baseline methods. Notably, ValuesRAG consistently outperforms all baselines across this validation data, achieving the highest accuracy despite relying exclusively on the values summaries. 

\begin{table}[htbp]
    \centering
    \begin{tabular}{@{}l@{\hspace{1.5em}}c@{\hspace{1.5em}}c@{}}
    \toprule
    \textbf{Methods} & \textbf{GPT-4o-mini} & \textbf{Gemini-1.5-flash} \\
    \midrule
    Zero-Shot & 0.6176 & 0.6041 \\
    Role-Assignment & \underline{0.6747} & \underline{0.6505} \\
    Few-Shot Learning & 0.6359 & 0.6086 \\
    Hybrid Method & 0.6670 & 0.6354 \\
    \midrule
    \textbf{Values Augmented} & \textbf{0.6894} & \textbf{0.6583} \\
    \bottomrule
    \end{tabular}
    \caption{Accuracy comparison between baseline methods and Values Augmented Generation method using the WVS validation set. \textbf{Bold text} indicates the best performance, 
    \underline{underlined text} indicates the second-best performance.}
    \label{tab:ablation_performance}
\end{table}

Compared to zero‐shot inference (0.6176 for GPT-4o-mini; 0.6041 for Gemini), Values-Only achieves 0.6894 with GPT-4o-mini and 0.6583 with Gemini, demonstrating structured value context provides substantially richer guidance than unconstrained prompts. Against the role-assignment (0.6747 for GPT-4o-mini; 0.6505 for Gemini), Values-Only attains higher scores as well, confirming that our generated value summaries capture more nuanced cultural patterns than static demographic labels. Similarly, Values-Only outperforms few-shot learning (0.6359 for GPT-4o-mini; 0.6086 for Gemini) and the hybrid method (0.6670 for GPT-4o-mini; 0.6354 for Gemini), further illustrating even without explicit examples or demographic anchors, our values-driven prompts produce more context-aligned predictions.

These results confirm the effectiveness and robustness of the values-augmented generation approach. ValuesRAG leverages structured values summaries to generate contextually rich and culturally aligned responses. Even without demographic augmentation, ValuesRAG achieves superior performance by dynamically capturing the underlying value patterns, demonstrating its ability to generalize across diverse cultural contexts without requiring predefined QA examples or demographic anchors. The results demonstrate the framework's scalability and adaptability, effectively mitigating biases and generating culturally coherent outputs with minimal dependence on external context.

% #################
\section{Case Study}
To further illustrate the effectiveness and practical utility of \emph{ValuesRAG}, we present two qualitative case studies derived from the \textsc{GSS} (United States) and \textsc{CGSS} (China) datasets. These case studies clearly demonstrate how our method dynamically forms a culturally and demographically relevant retrieval corpus, enhances values alignment, and consistently outperforms baseline approaches by providing richer, context-sensitive reasoning for the generation process.

\begin{figure}[htbp]
    \centering
    \includegraphics[width=0.85\linewidth]{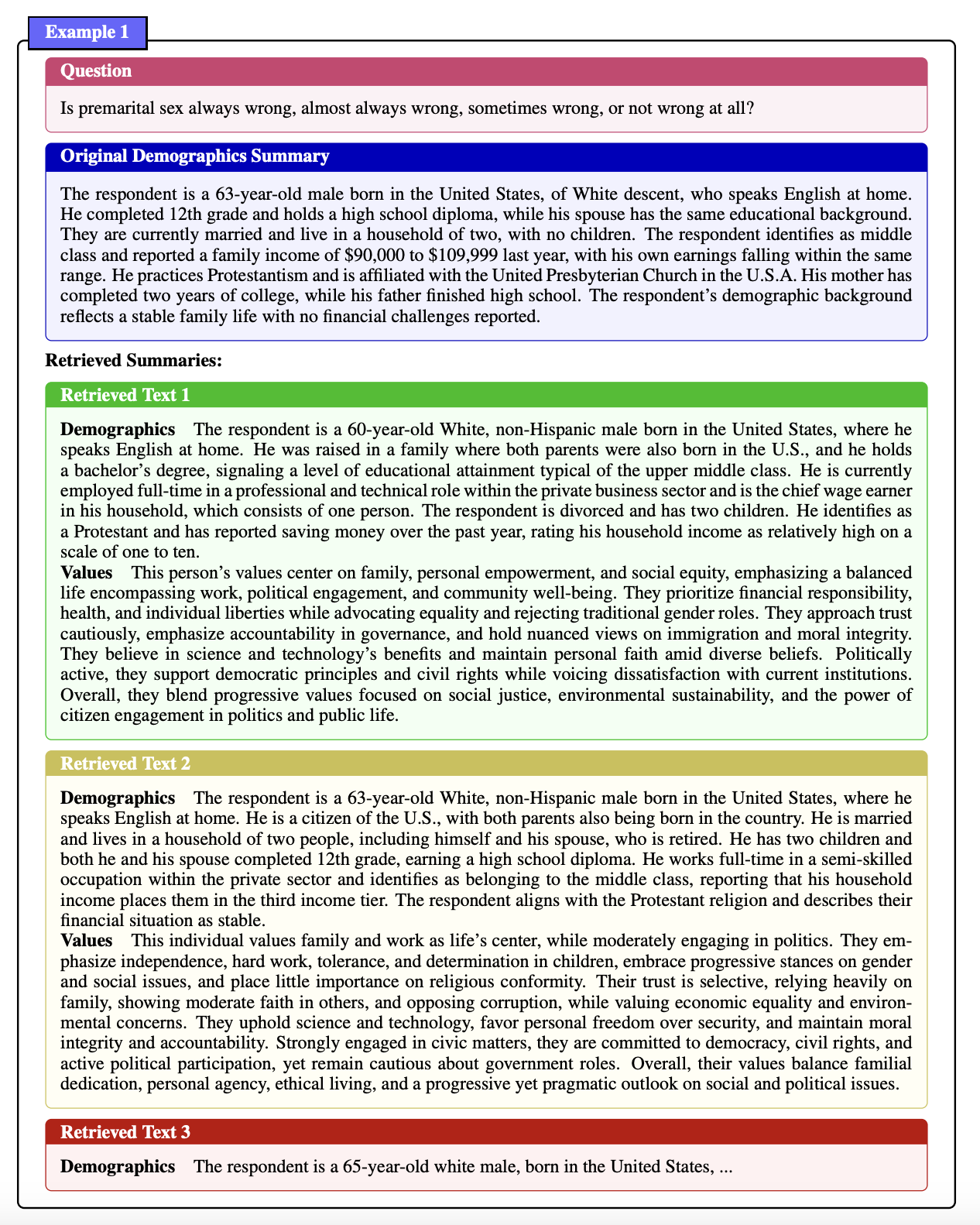}
    \caption{\textbf{Case Study (United States): Cultural Attitudes Toward Premarital Sex}. The original demographic profile (a 63-year-old Protestant male from the U.S.) is complemented by multiple retrieved summaries, enabling the LLM to reason with contextual sensitivity, avoiding stereotypes and enhancing values alignment.}
    \label{fig:qual_example1}
\end{figure}

\subsection{Case 1: Cultural Attitudes Toward Premarital Sex (United States)}
\label{case:us}
The first case (Figure~\ref{fig:qual_example1}) involves a 63-year-old middle-class Protestant male from the U.S., responding to the question: \textit{``Is premarital sex always wrong, almost always wrong, sometimes wrong, or not wrong at all?''} \emph{ValuesRAG} constructs the retrieval corpus by embedding demographic information and retrieving topically relevant profiles with similar backgrounds (e.g., age, ethnicity, education, and religious affiliation). Specifically, the top three retrieved profiles closely align demographically with the original individual yet introduce nuanced ideological variations. Each retrieved document not only reflects the target individual's demographic features but also articulates distinctive value orientations, highlighting the contextual breadth and depth of the retrieved corpus.

This carefully constructed retrieval corpus significantly enhances values alignment by providing multiple representative and nuanced perspectives, allowing the LLM to consider various contextually valid stances rather than reinforcing stereotypes typically found in static demographic-based methods. Compared to role-assignment or few-shot approaches, which struggle to represent this nuanced cultural diversity, ValuesRAG dynamically integrates subtle but critical distinctions across retrieved summaries. As a result, the generated response reflects realistic and culturally plausible reasoning that respects both traditional values and evolving societal norms, showcasing the superiority of ValuesRAG in handling culturally sensitive and complex value judgments.

\begin{figure}[htbp]
    \centering
    \includegraphics[width=0.85\linewidth]{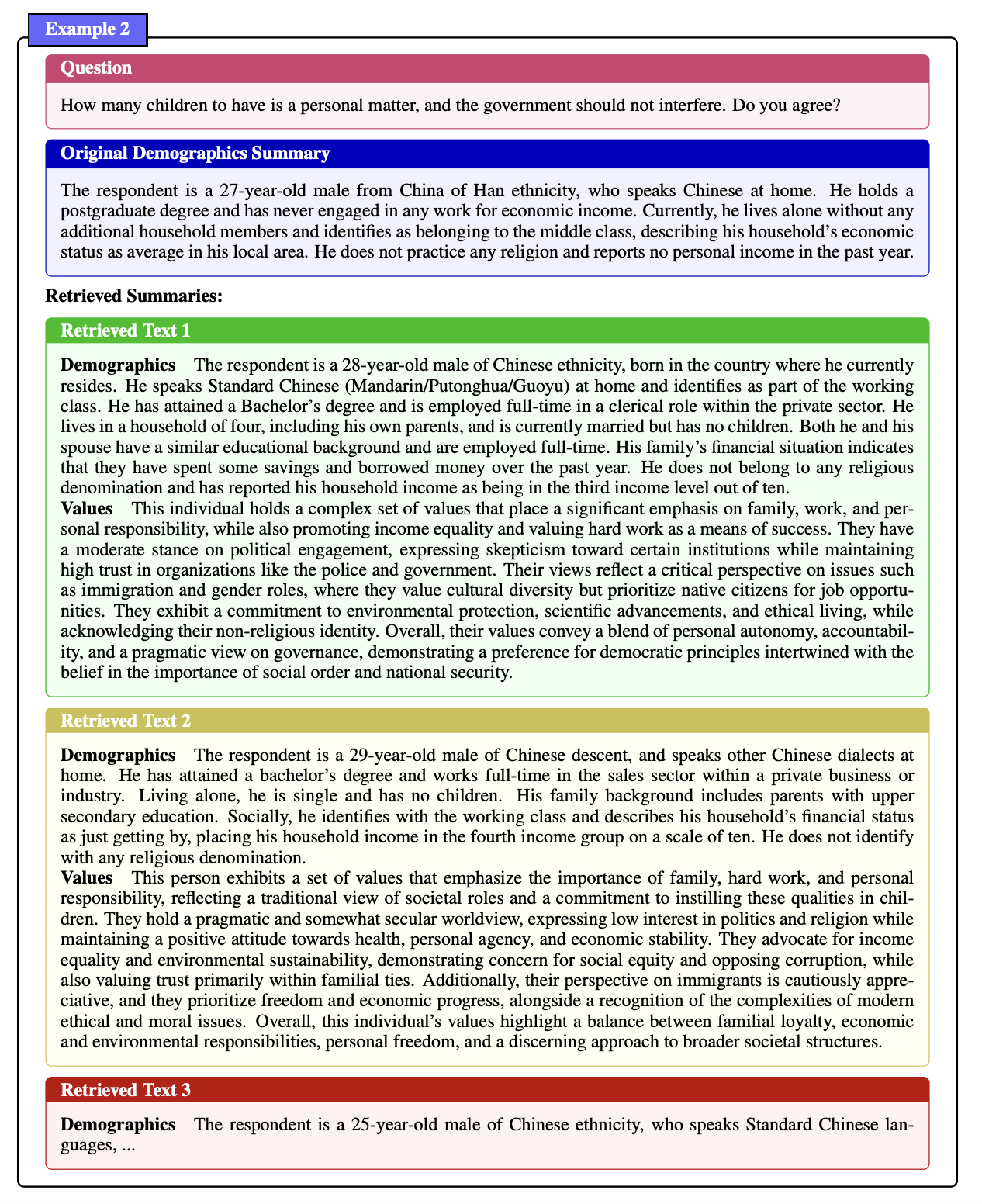}
    \caption{\textbf{Case Study (China): Family Planning Preferences.}  The retrieved summaries facilitate nuanced reasoning around personal autonomy and governmental roles, effectively capturing contemporary social shifts in attitudes toward family planning.}
    \label{fig:qual_example2}
\end{figure}

\subsection{Case 2: Family Planning Preferences (China)}
\label{case:china}
The second case (Figure~\ref{fig:qual_example2}) examines a 27-year-old secular male with postgraduate education from urban China addressing the question: \textit{``How many children to have is a personal matter, and the government should not interfere. Do you agree?''} In this scenario, \emph{ValuesRAG} retrieves profiles of individuals with closely matching demographics—such as age, education, urban residence, etc. Each retrieved summary reflects values related to personal agency, social responsibility, and family planning, with slight variations in economic status or professional roles. The constructed retrieval corpus is well-aligned with the original profile and relevant to the thematic context of the prompt.

By incorporating these diverse yet demographically coherent value profiles, \emph{ValuesRAG} enables the LLM to generate responses that account for a variety of socially grounded perspectives on family decision-making. In contrast to static prompting methods that may overlook such intra-cultural nuances, ValuesRAG captures shifts in value expression across population segments in a context-aware manner. This case shows how the method's dynamic retrieval mechanism enhances cultural alignment by offering broader coverage of relevant value orientations within the target demographic.

\section{Conclusion}
We propose \emph{ValuesRAG}, a novel framework designed to advance cultural values alignment through context-aware reasoning. In contrast to prior methods that rely on fixed demographic labels or limited few-shot prompting, \emph{ValuesRAG} dynamically retrieves and integrates contextually rich value summaries using adaptive retrieval, reranking, and In-Context Learning (ICL). Our contributions extend beyond the framework design to include a comprehensive evaluation across diverse, culturally and geographically representative test datasets. Extensive experiments demonstrate that \emph{ValuesRAG} consistently outperforms existing approaches, effectively capturing complex cultural nuances, reducing biases, and generating contextually aligned responses. Through its structured and adaptive design, \emph{ValuesRAG} bridges the gap between general-purpose LLMs and the demands of culturally sensitive applications, offering a scalable and robust solution for real-world deployment.

\section{Limitation}
Although our baseline comparisons indicate that ValuesRAG generally delivers superior performance compared to alternative methods, it does not always guarantee an exact match to an individual’s true values. Specifically, since our method relies on dynamically retrieved summaries from a predefined corpus (WVS), mismatches may occur when summaries derived from one dataset are applied to different or novel datasets. Future research should therefore explore more adaptive retrieval strategies capable of precisely aligning with novel datasets and investigate how integrating additional fine-tuning steps within retrieval-augmented generation frameworks could further enhance each agent’s contextual accuracy and generalizability across diverse value domains.

\section{Ethical Considerations}
While ValuesRAG effectively mitigates the common stereotype reinforcement seen in static methods, its reliance on demographic features inherently carries potential ethical risks related to profiling, bias, and fairness. If deployed in sensitive or high-stakes contexts such as public policy or persuasive communication, improper handling of demographic information could unintentionally perpetuate stereotypes or biases, adversely affecting vulnerable groups. We emphasize that our framework is intended not to reinforce specific demographic-driven stereotypes or biases, but rather to surface contextually relevant values explicitly for scrutiny.

Researchers and practitioners utilizing ValuesRAG must proactively examine and address ethical implications surrounding demographic-based retrieval. It is critical to integrate ValuesRAG into broader evaluative frameworks designed for continuous monitoring of potential societal impacts, ensuring its responsible deployment. Additionally, future work should explicitly investigate the ethical dimensions of demographic profiling within retrieval-augmented methods, further promoting fairness, accountability, and transparency in AI-driven cultural alignment tasks.

\appendix
\bibliography{aaai25}

\end{document}